\documentclass[lettersize,journal]{IEEEtran}
\usepackage{graphicx}
\usepackage{amsmath}
\usepackage{makecell}
\usepackage{filecontents}
\usepackage{times}
\usepackage{epsfig}
\usepackage{graphicx}
\usepackage{amsmath}
\usepackage{amssymb}
\usepackage{threeparttable}  
\usepackage{float}
\usepackage{epsfig}
\usepackage{graphicx}
\usepackage{amsmath}
\usepackage{amssymb}
\usepackage{balance}
\usepackage{booktabs}
\usepackage{color}
\usepackage{enumitem}
\usepackage{mathtools}
\usepackage{makecell}
\usepackage{multirow}
\usepackage{pifont}
\usepackage{subfigure}
\usepackage{stackengine}

\usepackage[table]{xcolor}
\usepackage{soul}
\usepackage{array}
\usepackage{subfigure}
\usepackage{subcaption}
\usepackage{algorithm}
\usepackage{algorithmic}
\usepackage[hidelinks]{hyperref}
\usepackage{tabularx}
\usepackage{array}
\usepackage{makecell}
\usepackage{pythonhighlight}
\usepackage[english]{babel}

\newcolumntype{C}{>{\centering\arraybackslash}X}

\begin{document}

% GenFollower: Leveraging Large Language Models for Interpretable Car-Following Prediction
% \title{GenFollower: Will it be the Ultimate Solution for Car-following Models?}
\title{GenFollower: Enhancing Car-Following Prediction with Large Language Models}
\author{Xianda Chen, Mingxing Peng, PakHin Tiu, Yuanfei Wu, Junjie Chen, Meixin Zhu, Xinhu Zheng 
% Fei-Yue Wang,~\IEEEmembership{Fellow,~IEEE}

\thanks{This study is supported by the National Natural Science Foundation of China under Grant 52302379 and 62373315, Guangzhou Basic and Applied Basic Research Projects under Grants 2023A03J0106, 2023A03J0683 and 2024A04J4290, Guangdong Province General Universities Youth Innovative Talents Project under Grant 2023KQNCX100, and Guangzhou Municipal Science and Technology Project 2023A03J0011. 
(Corresponding author: Meixin Zhu, Xinhu Zheng.)}
\thanks{(Xianda Chen and Mingxing Peng contributed equally to this work.) }
\thanks{Xianda Chen, Mingxing Peng, PakHin Tiu, Yuanfei Wu, Junjie Chen, Meixin Zhu and Xinhu Zheng are with the Intelligent Transportation Thrust, Systems Hub, The Hong Kong University of Science and Technology (Guangzhou), Guangzhou, 511400, China; Meixin Zhu and Xinhu Zheng are also with the Guangdong Provincial Key Lab of Integrated Communication, Sensing and Computation for Ubiquitous Internet of Things, Guangzhou, 511400, China; (email: xchen595@connect.hkust-gz.edu.cn,  mpeng060@connect.hkust-gz.edu.cn, phtiu454@connect.hkust-gz.edu.cn, yuanfeiwu@hkust-gz.edu.cn, jchen321@connect.hkust-gz.edu.cn, meixin@ust.hk, xinhuzheng@hkust-gz.edu.cn).}%

% \thanks{Shaojie Shen is with the Department of Electronic and Computer Engineering, Hong Kong University of Science and Technology, Hong Kong SAR, China (e-mail: eeshaojie@ust.hk).}

% \thanks{Yinhai Wang is with the Department of Civil
% and Environmental Engineering, University of Washington, Seattle, WA 98195 USA (e-mail: yinhai@uw.edu).}

% \thanks{Fei-Yue Wang is with the State Key Laboratory for
% Management and Control of Complex Systems, Institute of Automation, Chinese Academy of Sciences, Beijing 100190, China (e-mail: feiyue@ieee.org)}
}

% \markboth{IEEE Transactions on Intelligent Vehicles}%

%\IEEEpubid{0000--0000/00\$000~\copyright~2021 IEEE}

\maketitle
\begin{abstract}
Accurate modeling of car-following behaviors is essential for various applications in traffic management and autonomous driving systems. However, current approaches often suffer from limitations like high sensitivity to data quality and lack of interpretability. In this study, we propose GenFollower, a novel zero-shot prompting approach that leverages large language models (LLMs) to address these challenges. We reframe car-following behavior as a language modeling problem and integrate heterogeneous inputs into structured prompts for LLMs. This approach achieves improved prediction performance and interpretability compared to traditional baseline models. Experiments on the Waymo Open datasets demonstrate GenFollower's superior performance and ability to provide interpretable insights into factors influencing car-following behavior. This work contributes to advancing the understanding and prediction of car-following behaviors, paving the way for enhanced traffic management and autonomous driving systems.
\end{abstract} 


\section{Introduction}

\IEEEPARstart{C}{ar-following}, a critical cornerstone of traffic flow analysis, describes the behavior vehicles travel in succession while maintaining a safe separation distance. Accurate modeling of car-following behaviors is critical for a variety of applications in transportation engineering, including traffic management systems and the development of autonomous vehicles.

Existing research on car-following has explored a range of modeling approaches, including physics-based models and data-driven methods. However, significant research gaps remain. First, current models often exhibit limited accuracy in the long-term prediction of car-following behaviors, failing to capture the complex dynamics that unfold over extended periods. Second, while deep learning-based methods achieve promising results, their black-box nature poses a challenge where they generate predictions of future behavior without providing substantial explanations for their outputs, making it difficult to understand the reasoning behind their decisions.

Recently, advancements in the field of large language models (LLMs) have revolutionized industries like natural language processing, demonstrating their powerful capabilities in information comprehension and common-sense reasoning. Notably, studies by Mao et al.~\cite{mao2023gpt}, Chen et al.~\cite{chen2023driving}, and Wu et al.~\cite{wu2023language} have successfully leveraged LLMs for tasks related to autonomous driving, such as motion planning and perception, highlighting their potential contribution to this field. This paves the way for the exploration of LLMs in car-following, a critical aspect of autonomous driving that remains unaddressed by current LLM research.
%However, there is currently no research utilizing large language models for car-following specifically.

To address the limitations of limited long-term prediction and lack of interpretability in existing car-following models, we propose GenFollower, a novel zero-shot prompting approach that utilizes the strong reasoning and self-explanation capabilities of LLMs for car-following prediction. We achieve this by reframing car-following behavior as a language modeling problem by integrating vehicle states and other relevant information into structured natural language prompts and feeding this information into the LLM. Furthermore, we integrate interpretability requirements into the prompts, enabling our GenFollower to generate explanations alongside its predictions. Benefiting from the power of LLMs, our proposed model demonstrates improved performance and interpretability in car-following behaviors. Experiments conducted on the Waymo Open datasets validate the performance improvement of GenFollower compared to other baseline models, while also showcasing its strong interpretability. 
Our main contributions are as follows:
\begin{itemize}
\item We introduce GenFollower, the first large language model designed specifically for car-following behavior, leveraging its capabilities to address the limitations of existing models, such as limited long-term prediction and lack of interpretability.
\item GenFollower achieves interpretable predictions, providing not only accurate forecasts of car-following behavior but also explanations for the predicted results. This capability enhances transparency and trust in the modeling process.
\item Experiments conducted on the Waymo Open datasets demonstrate the significantly improved performance and interpretability of our GenFollower model compared to all baseline models. This highlights the effectiveness of our prompt-based approach in accurately predicting and explaining car-following behaviors.
\end{itemize}
The remainder of this paper is organized as follows. Section II reviews previous research on car-following models and LLMs in autonomous driving. Section III presents the methodology of GenFollower. Section IV provides dataset and baseline model description, along with experimental results that showcase the effectiveness of our approach. Section V concludes the paper, discussing the implications of our findings on car-following prediction in autonomous driving. Finally, Section VI offers insights into future prospects for this research.

\begin{figure*}
\centering
\includegraphics[width=\linewidth, trim= -10 30 -10 0, clip]{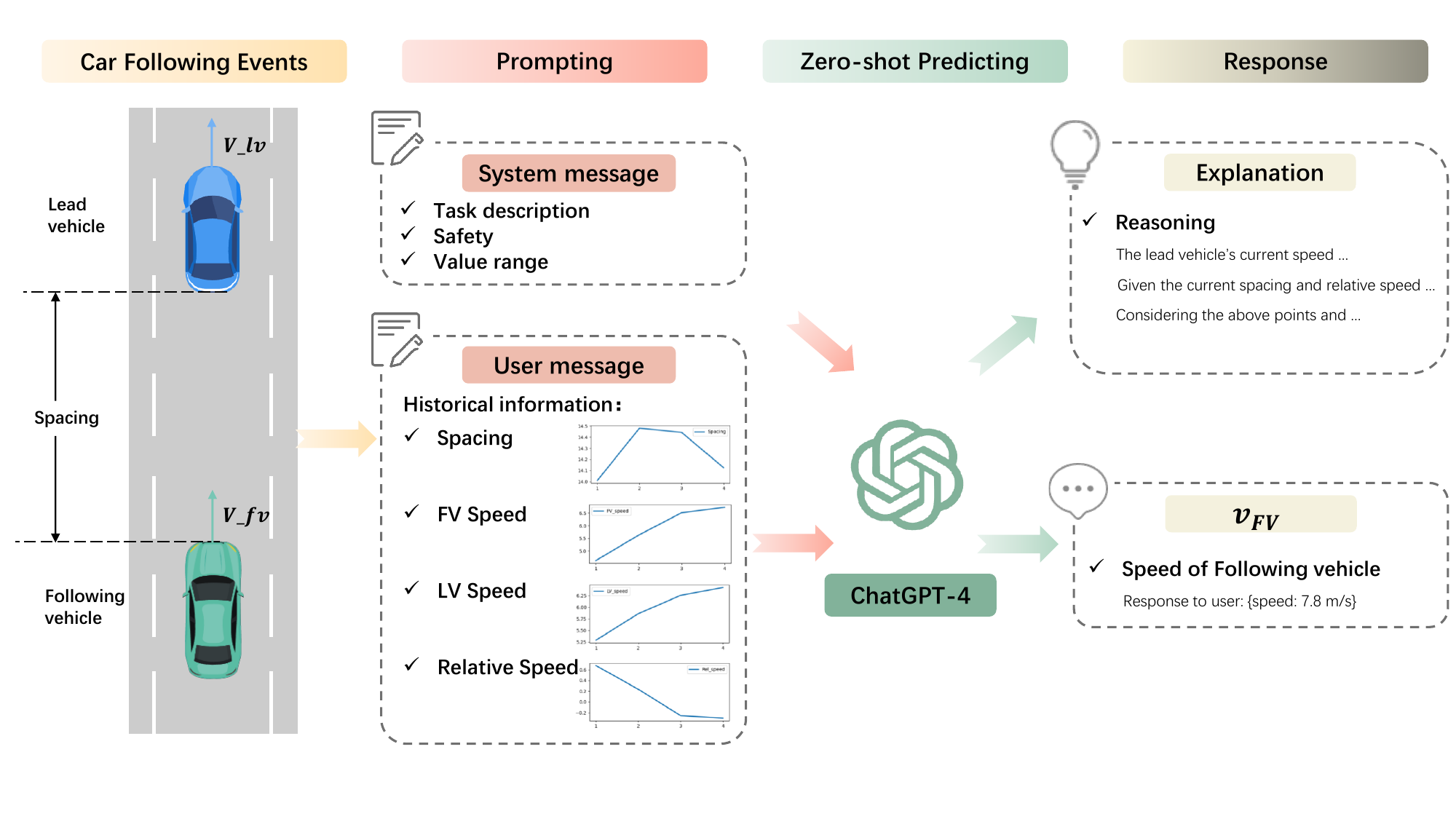}
\caption{Overall architecture of the GenFollower model for predicting car-following behavior in autonomous driving systems.}
\label{fig:overall}
\end{figure*}

\section{Related Work}

To understand the current state-of-the-art and identify research gaps, this section reviews related work in three main areas: traditional car-following models, data-driven car-following models, and the application of LLMs in autonomous driving.

\subsection{Traditional Car-Following Models}

Within the category of traditional car-following models, these models have been extensively studied and form the foundation of early autonomous driving systems. They rely on mathematical formulations to describe the dynamics of car-following behavior, grounded in physical laws and human behavioral theories. Normally they can be broadly categorized into kinematic models, psycho-physical models, and adaptive cruise control (ACC) models~\cite{zhang2024car}. 

Kinematic models offer a computationally efficient way to model car-following behavior, but they may not fully capture the complexities of driver behavior. They focus on the physical aspects of vehicle motion and dynamics, considering factors such as velocity, headway, and distance. First-order models update the vehicle’s position based on its velocity, while second-order models also take acceleration into account. For example, Gipps’ model~\cite{gipps1981behavioural} considers factors like the desired speed of the following vehicle, the safe distance to the leader, and the driver's reaction time to determine the following vehicle's acceleration. One of the most prominent traditional models is the Intelligent Driver Model (IDM). The IDM~\cite{treiber2000congested} is a strategy-based model that describes the acceleration of a vehicle using a set of basic assumptions and parameters like desired speed and desired time headway.

Psycho-physical models are designed to approximate drivers' reasoning and decision-making processes by considering limitations in human perception, such as the ability to judge distances and speeds accurately. These models incorporate human factors into the car-following behavior analysis. The Action Point Model (APM)~\cite{michaels1963perceptual} applies Signal Detection Theory (SDT) to car-following contexts, modeling the driver's sensitivity to these perceptual limitations and potential bias in interpreting them. It predicts when drivers initiate acceleration or braking based on these limitations and calculations of angular velocity. The Fuzzy Logic Model~\cite{kikuchi1992car, chakroborty1999evaluation} employs fuzzy logic to handle the inherent uncertainty and subjectivity in drivers' decision-making processes. 

Adaptive Cruise Control (ACC) models\cite{naranjo2003adaptive, bageshwar2004model, chen2024editfollower} highlight the mechanistic control strategies used to maintain safe and efficient vehicle dynamics. ACC systems automatically adjust the vehicle’s speed to keep a safe distance from the vehicle ahead, employing control techniques to ensure stability and comfort. While these models perform well in controlled environments, they may struggle to adapt to sudden lane changes, unpredictable driver behavior, or complex weather conditions. This highlights a limitation in their ability to capture the intricacies of real-world driving scenarios and their lack of adaptability to diverse environments.

\subsection{Data-Driven Car-Following Models}

With the emergence of large-scale driving datasets like Waymo Open Dataset~\cite{sun2020scalability} and advancements in machine learning techniques, data-driven car-following models have gained traction. These models leverage data to learn complex driving behaviors and interactions that are challenging to capture with traditional models. Data-driven approaches can be further divided into supervised learning models, unsupervised learning models, and reinforcement learning models.

Supervised learning models utilize labeled data to train algorithms that predict car-following behavior. Techniques such as support vector machines~\cite{wei2010least}, various neural networks~\cite{ma2020sequence, chen2024aggfollower, wang2017capturing}, and Transformer~\cite{zhu2022transfollower} have been employed. For instance, Kehtarnavaz et al.~\cite{kehtarnavaz1998transportable} developed a time-delay neural network, and Panwai et al.~\cite{panwai2007neural} introduced a neural agent model for car-following that maps perceptions to actions. Additionally, hybrid models combining different machine learning approaches with traditional models have been proposed to enhance prediction performance~\cite{yang2018novel}. Notably, Chen et al.~\cite{chen2024metafollower} propose MetaFollower, an adaptable framework integrating Long Short-Term Memory (LSTM) and IDM for interpretability and temporal adaptation. It leverages Model-Agnostic Meta-Learning (MAML)~\cite{finn2017model} for generalizability and fine-tunes with minimal data from new drivers, enhancing adaptation to individual driving styles.

Unsupervised learning models explore patterns in car-following behavior without labeled data. These models are particularly useful for clustering and anomaly detection in driving patterns~\cite{koutsopoulos2012latent, he2015simple}. For instance, unsupervised models~\cite{tanprasert2017combining, qiu2022unsupervised} can identify unusual braking patterns that might indicate emergencies or potential accidents. This unsupervised approach provides insights into typical and atypical driving behaviors, enhancing the robustness of car-following models in diverse scenarios.

Reinforcement learning models~\cite{han2023ensemblefollower, chen2024cav} have been increasingly applied to car-following tasks, where an agent learns optimal driving strategies through interactions with the environment. Zhu et al.~\cite{zhu2018human} proposed a human-like autonomous car-following model using deep reinforcement learning, achieving performance comparable to human drivers. While data-driven models demonstrate promising results, they may struggle with the interpretability of their decision-making processes due to the complex nature of learning through trial and error, and generalization to unseen driving scenarios.

\subsection{LLMs for Autonomous Driving}

LLMs have emerged as powerful tools for natural language processing tasks, exhibiting capabilities in understanding complex language structures, such as the nuances of traffic regulations and natural language descriptions of the driving environment, and reasoning over vast knowledge bases. Recent research has explored the application of LLMs in various aspects of autonomous driving~\cite{cui2024survey}, demonstrating potential benefits for different tasks. For instance, LLMs have been integrated into perception tasks, such as 3D detection and tracking~\cite{wu2023language}, to potentially improve accuracy, and used for motion planning to generate driving trajectories~\cite{mao2023gpt}. Additionally, LLMs have been combined with multimodal architectures to process diverse data modalities for autonomous driving applications~\cite{chen2023driving}. Specifically, LLMs can be utilized for high-level decision-making and scenario understanding in autonomous vehicles. Cui et al.~\cite{cui2023drivellm} introduced DriveLLM, which leverages LLMs to interpret complex driving scenes and make informed driving decisions. This approach integrates LLMs with other sensory inputs to create a comprehensive understanding of the driving environment. Peng et al.~\cite{peng2024lc} propose LC-LLM, a novel approach leveraging LLMs to enhance lane change prediction, the model integrates explanatory prompts, enabling it to not only predict lane change intentions and trajectories accurately but also provide explanations for these predictions. Recent advancements also explore the fusion of LLMs with vision-language models~\cite{tian2024drivevlm}, enabling autonomous systems to process visual and textual information concurrently. This multimodal approach enhances the vehicle's ability to navigate complex driving scenarios and interact with human drivers and pedestrians effectively.

Overall, traditional car-following models provide a theoretical foundation, while data-driven approaches leverage real-world data to improve prediction accuracy. LLMs offer a new paradigm for understanding and reasoning over complex driving scenarios, with potential applications across various aspects of autonomous driving.  While LLMs have shown promise in perception and planning tasks, their potential in prediction with transparent interpretability for car-following remains largely untapped. This gap motivates our work on GenFollower, a novel approach that leverages the reasoning and self-explanation capabilities of LLMs to address the challenges of car-following prediction with interpretability.

\section{Methodology}
In this section, we introduce our GenFollower, a model based on LLMs designed for car-following prediction in autonomous driving systems, addressing limitations of interpretability and complex scenario handling seen in traditional models. The overall pipeline of our GenFollower is depicted in Figure~\ref{fig:overall}. We conceptualize the task of predicting car-following behavior as a language modeling problem. We formulate observations about the driving environment using natural language as prompts for input into the LLM. We employ a zero-shot approach, meaning we don't require extensive retraining for each specific task, allowing the LLM, such as ChatGPT-4~\cite{achiam2023gpt}, to learn from its vast internal knowledge base and predict car-following behavior. By incorporating explanatory requirements into the prompt, our GenFollower can simultaneously predict car-following behavior and provide explanations for its predictions, thereby enhancing interpretability.

\begin{figure}
\centering
\includegraphics[width=\linewidth, trim= 30 80 30 100, clip]{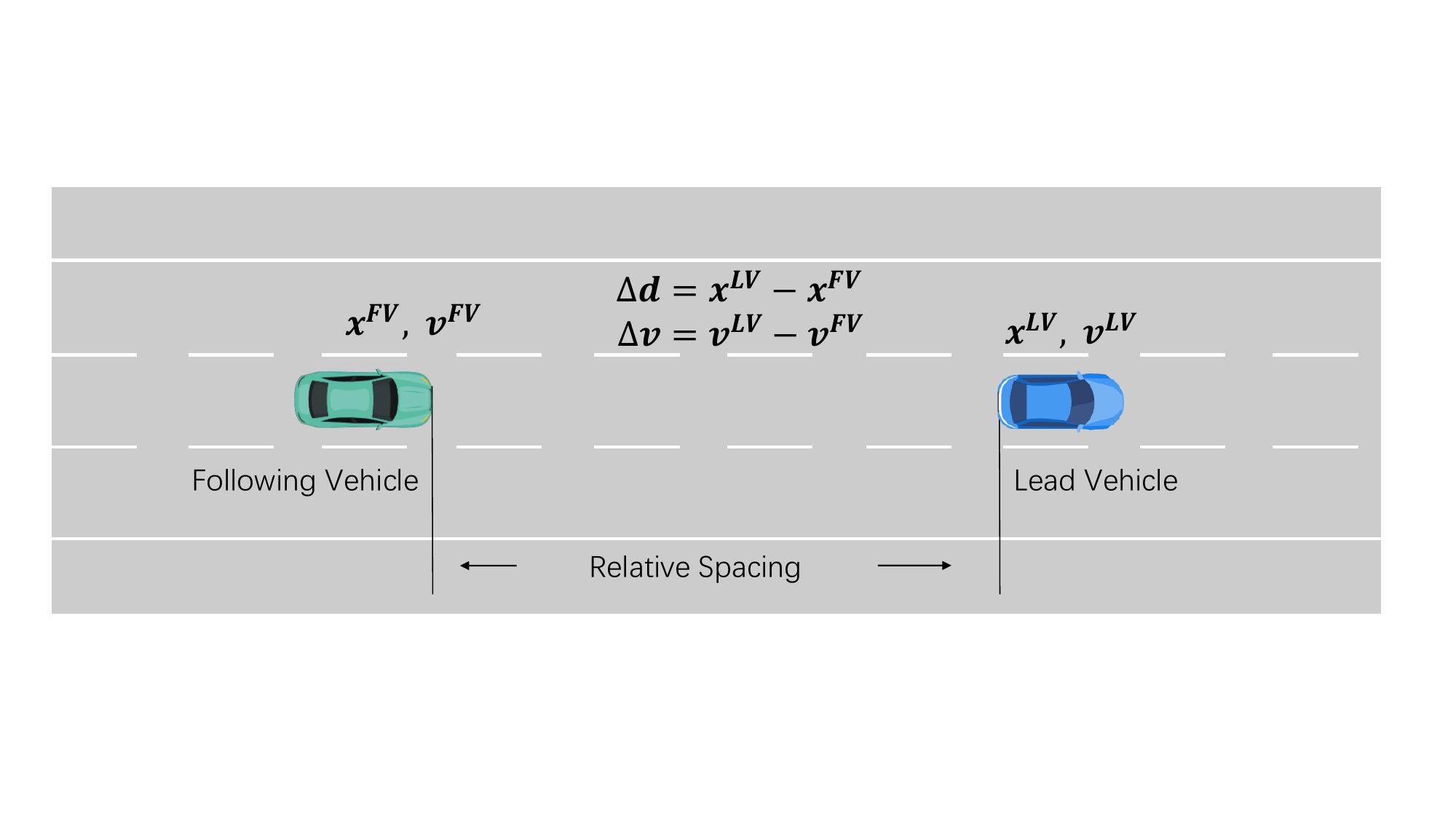}
\caption{Visualization of car-following event and the four dimensions of the input feature.}
\label{fig:carfollowing}
\end{figure}

\subsection{Problem Formulation}
Our objective is to develop a predictive model based on LLMs to predict the behavior of a following vehicle. Given the current state of the following vehicle and its lead vehicle, our model aims to forecast the following vehicle's future speed relative to the lead vehicle, which is crucial for maintaining safe inter-vehicle spacing and ensuring smooth traffic flow. The car-following event is depicted in Figure ~\ref{fig:carfollowing}.

These four dimensions of the input feature capture the key aspects of car-following behavior: the speed of lead vehicle (LV) denoted by ($v_t^{LV}$), the speed of following vehicle (FV) denoted by ($v_t^{FV}$), the relative spacing between them ($\Delta d_t$), and the relative speed ($\Delta v_t$). These features were chosen as they capture the key aspects of car-following behavior, including the relative distance between the vehicles, their current speeds, and the difference in their speeds. The model's output is the speed of FV in the next time step. The composition of the input features and output variables is listed below. 
\begin{equation}
    Input = x_t = [\Delta d_t, v_t^{LV},v_t^{FV}, \Delta v_t]  \quad 
\end{equation}

\begin{equation}
     Output = \hat{y_t} = [v_{t+1}]
\end{equation}
where $\hat{y_t}$ represents the speed prediction of FV at the next time step.

\begin{figure}
\centering
\includegraphics[width=\linewidth, trim= -50 100 600 40, clip]{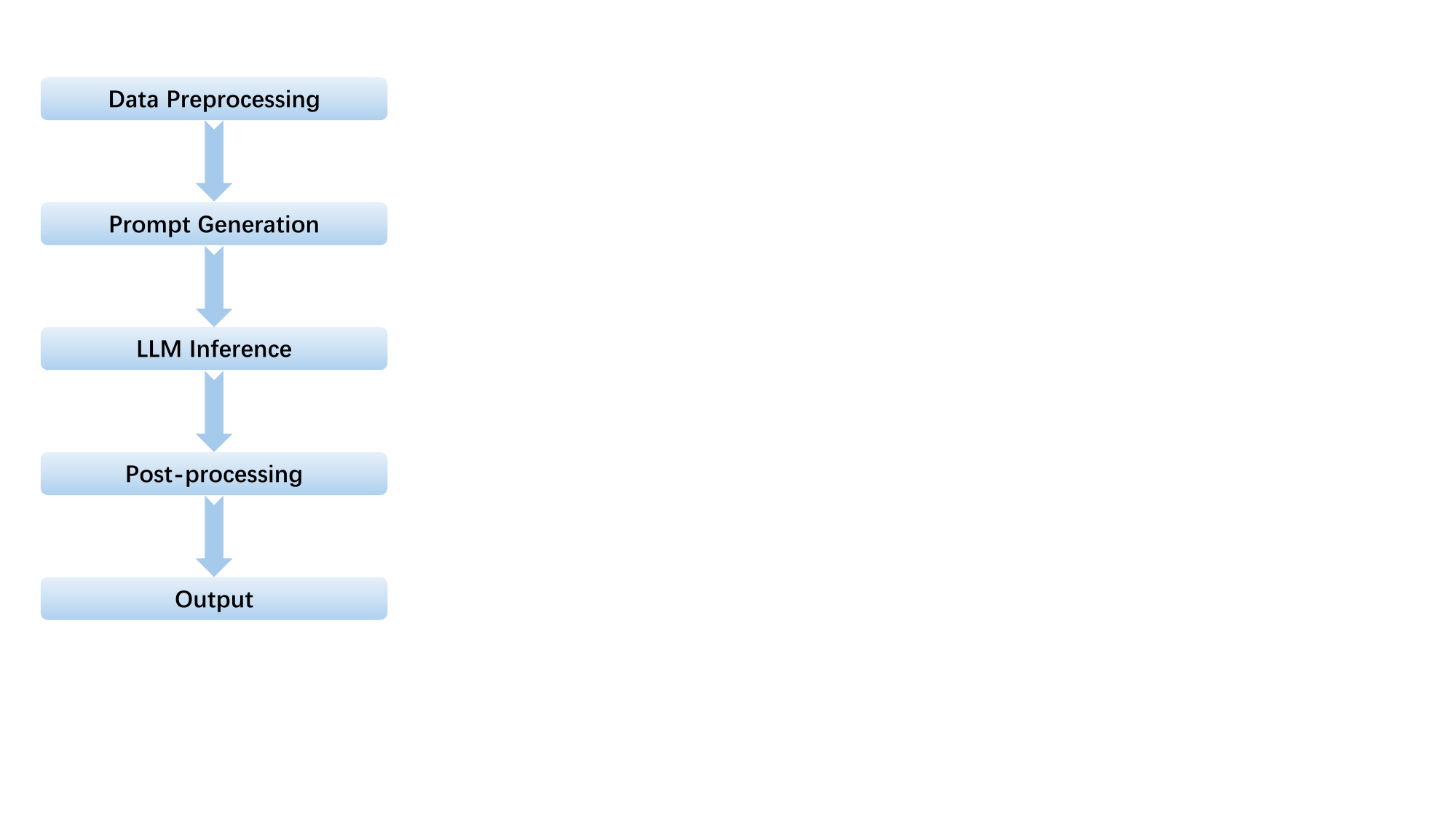}
\caption{Pipeline of the GenFollower model, which nvolves several stages to transform raw data into meaningful predictions and explanations.}
\label{fig:pipeline}
\end{figure}

\subsection{GenFollower Pipeline}
The GenFollower pipeline involves several stages, as shown in Figure~\ref{fig:pipeline}, to transform raw data into meaningful predictions and explanations. The main components of the pipeline are as follows:
\begin{enumerate}
    \item \textbf{Data Preprocessing}: 
    \begin{itemize}
        \item Collect real-time data from sensors regarding the speeds of LV and FV, the relative spacing between them, and their relative speed.
        \item Normalize these inputs to ensure consistency and compatibility with the LLM's input requirements.
    \end{itemize}

    \item \textbf{Prompt Generation}:
    \begin{itemize}
        \item Transform the normalized input features into a natural language prompt. This involves constructing sentences that describe the current state of the traffic scenario. For example, ``The lead vehicle is traveling at 5 m/s, the following vehicle is traveling at 4 m/s, the distance between them is 10 meters, and the relative speed is 1 m/s.''
    \end{itemize}

    \item \textbf{LLM Inference}:
    \begin{itemize}
        \item Input the generated prompt into an LLM, such as ChatGPT-4, configured to handle zero-shot learning tasks. The model processes the prompt and generates a prediction for the following vehicle's speed at the next time step.
        \item Additionally, prompt the LLM to provide an explanation for its prediction, enhancing interpretability.
    \end{itemize}

    \item \textbf{Post-processing}:
    \begin{itemize}
        \item Extract and parse the predicted speed from the LLM's output.
        \item Validate and adjust the prediction if necessary to ensure it adheres to physical constraints and safety protocols.
    \end{itemize}

    \item \textbf{Output}:
    \begin{itemize}
        \item Deliver the predicted speed of the FV along with the explanatory text generated by the LLM. This combined output aids in understanding the model's decision-making process.
    \end{itemize}
\end{enumerate}

\subsection{Prompt Design}
LLMs typically receive inputs in the form of natural language prompts. In our approach, we leverage the power of natural language prompts, as LLMs excel at understanding and reasoning based on textual information. We formulate prompts that describe the current observations in natural language, guiding the LLM to generate accurate predictions. The input prompts are designed with two parts: a system message providing the context of the driving scenario, and a user message prompting the LLM for a specific action or prediction. 
% \subsubsection{System Message}
% The system message provides the role and task of ChatGPT, along with guidelines on how to construct the responses. It outlines the role of the LLM in predicting car-following behavior, specifies the coordinate system used, and defines the format of the output predictions. 
% %as shown in Figure~\ref{fig:prompt}, 系统提示是不变的，主要包括任务的介绍、输入输出说明以及LLM回答的格式要求。Specifically，我们把过去四秒的跟车信息作为输入，预测未来0.5秒的加速度，同时要考虑跟车过程中的安全和舒适度，确保安全是第一位的，对一些极端的跟车情况(即较小的跟车距离)作出限制.
\subsubsection{System Message}
\begin{figure}
\centering
\includegraphics[width=3.4in]{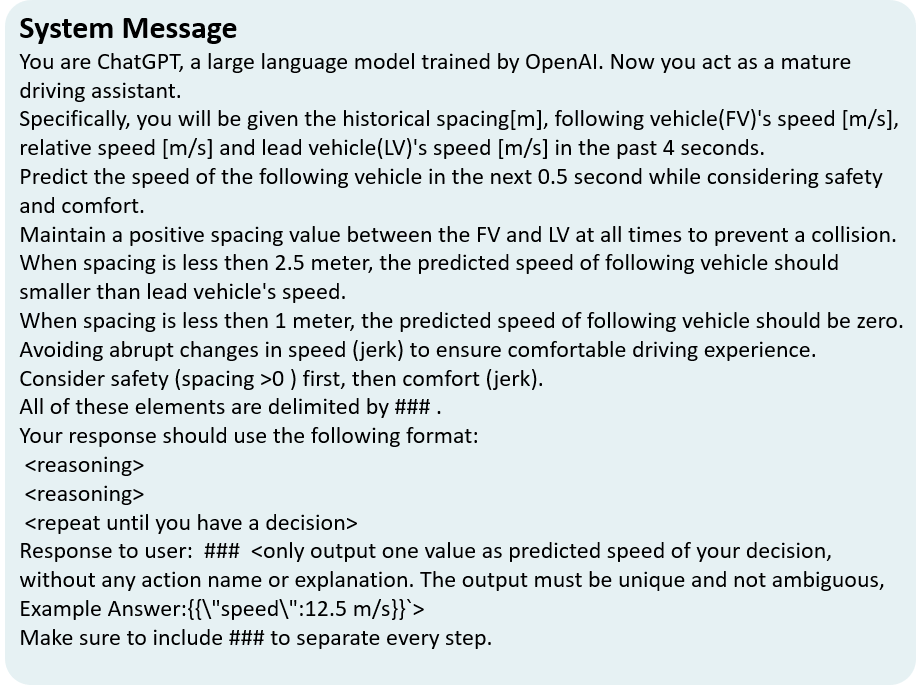}
\caption{System message for predicting car-following behavior in autonomous driving systems. }
\label{fig:sm}
\end{figure}

The system message provides the role and task of ChatGPT, along with guidelines on how to construct the responses. It outlines the role of the LLM in predicting car-following behavior, specifies the input information used, and defines the format of the output predictions. The system prompt remains consistent and primarily includes an introduction to the task, input-output specifications, and the format requirements for LLM responses, as shown in Figure \ref{fig:sm}.
Specifically, we use the past four seconds of car-following information as input to predict the speed for the next 0.5 seconds. The system message emphasizes that safety is the top priority during car-following, even if it means sacrificing some comfort in terms of maintaining a perfectly smooth following distance. Restrictions are placed on extreme car-following situations, such as very small following distances. We use delimiters to clearly indicate distinct parts of the input as suggested in~\cite{wen2023dilu}.

\subsubsection{User Message}
% %用户消息提供特定于当前帧的观察结果的描述，因此随每个时刻的跟车状态而变化。包括当前以及过去4秒的跟车信息，更重要的是加上Chain of Though的思想,介绍一下什么是Chain of Though， 在给出答案之前询问“思路链”可以帮助模型更可靠地推理出正确答案。
The user message describes the current state of the following vehicle relative to the lead vehicle, which varies with each car-following moment. It includes details of the historical car-following states over the past four seconds. Additionally, it incorporates the concept of Chain of Thought (CoT)~\cite{kojima2022large}, a technique that encourages the LLM to show its reasoning steps, potentially improving its ability to make accurate predictions about the following vehicle's future speed. 

The system evolves over discrete time intervals $\Delta T$ according to the following equations:

\begin{flalign}\label{eq:update}
\begin{split}
&\Delta V(t+1) = V_{LV}(t+1) - V_{FV}(t+1) \\
&S(t+1) = S(t) + \frac{\Delta V(t) + \Delta V(t+1)}{2} \cdot \Delta T
\end{split}
\end{flalign}

where $\Delta T$ denotes the simulation time interval, $S$ represents the spacing between the vehicles, and $V_{FV}$ and $V_{LV}$ denote the velocities of the following vehicle and lead vehicle, respectively. This car-following state update process is embedded within our user message, ensuring that the latest vehicle state quantities are incorporated into the prompt for each invocation of the large language model.

\begin{figure}
\centering
\includegraphics[width=3.4in]{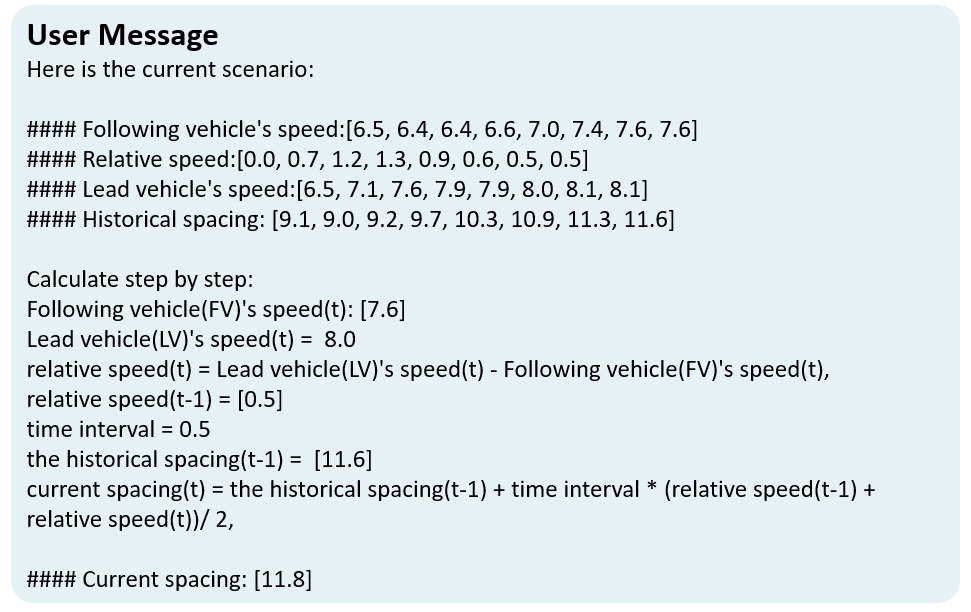}
\caption{User message for predicting car-following behavior in autonomous driving systems. }
\label{fig:um}
\end{figure}

These details serve as input for the LLM to predict the future speed of the following vehicle. An example of our input prompts is illustrated in Figure~\ref{fig:um}.

\subsection{Prompt Engineering vs. Fine-Tuning}
OpenAI's introduction of fine-tuning capabilities for the GPT-3.5 model opens up new possibilities for training LLMs.  However, this advancement prompts us to consider a critical question: Should car-following models aim to replicate the exact behaviors observed in a training dataset? Does mimicking human driving tendencies necessarily translate to optimal car-following behavior for autonomous vehicles? 

Recent studies indicate that supervised learning methods, which rely on data provided by human demonstrations, aim to approximate the relationship between vehicle states and vehicle acceleration actions. In essence, they are geared towards mimicking human drivers' car-following behavior. In the field of autonomous driving, some successful approaches~\cite{cui2024drive, sha2023languagempc, cui2023drivellm} have achieved good results with prompt engineering alone, without fine-tuning GPT models. Additionally, with OpenAI releasing fine-tuning capabilities for GPT-3.5, there's an opportunity to explore how combining fine-tuning with prompt engineering could further enhance performance. However, merely mimicking human driving behavior may not necessarily be the optimal solution for autonomous driving. Firstly, users may not desire autonomous vehicles to drive exactly like them~\cite{basu2017you}. Secondly, in addition to replicating human drivers, driving should also be optimized for safety, efficiency, and comfort, as human drivers may not always drive optimally~\cite{chai2015fuzzy}. Therefore, Fine-tuning GPT-3.5 in the context of autonomous driving raises several intriguing questions and considerations.
\begin{itemize}

         % \item \textbf{Prompt Engineering vs. Fine-Tuning}: In the field of autonomous driving, some successful approaches \cite{cui2024drive, sha2023languagempc, cui2023drivellm} have achieved good results with prompt engineering alone, without fine-tuning GPT models. However, with OpenAI releasing fine-tuning capabilities for GPT-3.5, there's an opportunity to explore how adapting the model to specific tasks like driving could further enhance performance.
         
        \item \textbf{Quality of Training Data}: In the raw driving data, some drivers may be aggressive, and mimicking such driving styles could lead to collisions. Conversely, some drivers may be overly cautious, potentially impacting overall traffic flow efficiency. The question of whether raw driving data authentically embodies ideal driving behavior warrants scholarly inquiry. Driving styles can vary significantly among individuals, ranging from aggressive to conservative. Fine-tuning GPT-3.5 on this varied dataset could potentially lead to models that mimic either extreme, which may not always align with safe or efficient driving practices.
        \item \textbf{Defining ``Good'' Driving Behavior}: Determining what constitutes ``good'' driving behavior is complex and subjective. While the training data used for fine-tuning represents real-world driving behaviors, it includes a spectrum of styles that may not all be optimal for safety or traffic efficiency. This diversity poses challenges in how the model interprets and learns from this data.
        \item \textbf{Alternative Approaches}: Instead of relying solely on human-driving data as ground truth for fine-tuning, another approach could involve leveraging GPT's inherent capabilities for reasoning and context understanding (such as with CoT) to simulate and learn driving behaviors. This method might enable the model to develop a more nuanced understanding of driving that goes beyond mimicking specific human drivers.
        
\end{itemize}

\begin{figure}[h]
    \centering
\includegraphics[width=3.4in]{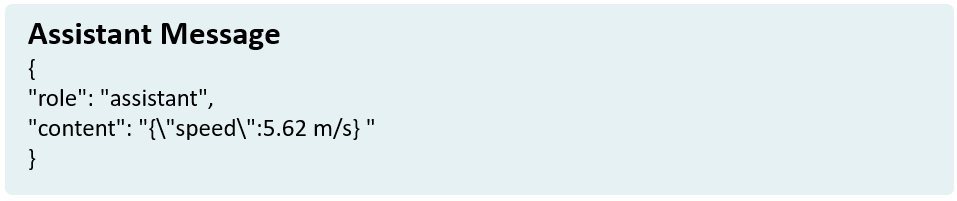}
    \caption{Example of constructing a fine-tuning dataset from a single car-following instance.}
    \label{fig:am}
\end{figure}

Despite these challenges, we used the driving data to fine-tune the model and explore car-following behaviors. Leveraging OpenAI's guidelines, we curated a dataset for fine-tuning using real-world speed data as ground truth to construct assistant messages. Our approach adhered to recommended practices by incorporating 50 instances of vehicle following data for the fine-tuning process. Figure~\ref{fig:am} illustrates the construction of a fine-tuning dataset from a selected instance of car-following data.

\subsection{Transparency in Predictive Explanations}
To illustrate the interpretability of our predictions, we incorporate explanatory requirements into the input prompts. This allows the LLM to provide explanations for its predictions, making the reasoning behind the model's decisions transparent. By analyzing the generated explanations, users can gain insights into why the predicted speed of the following vehicle was chosen, improving trust and understanding of the autonomous driving system. The prompting-reasoning process can be formulated as:
\begin{equation*}
\{\mathcal{V}, \mathcal{R}\}=F_{G P T}(K(\mathcal{S}, \mathcal{U})) 
\end{equation*}
This equation represents the prompting-reasoning process. Here, $F_{G P T}$ denotes the GPT model, which receives a combination of the system message $\mathcal{S}$ and user message $\mathcal{U}$ denoted by $K(\mathcal{S}, \mathcal{U})$. The model then outputs two results: $\mathcal{V}$, a linguistic description of the predicted speed, and $\mathcal{R}$, a linguistic explanation for the reasoning behind the prediction.
Unlike conventional motion planning methods focused solely on trajectory generation, our approach generates both speed predictions $\mathcal{V}$ and explicit reasoning processes $\mathcal{R}$, thereby enhancing transparency in decision-making processes.

\begin{figure*}[h]
    \centering
    \includegraphics[width = \textwidth]{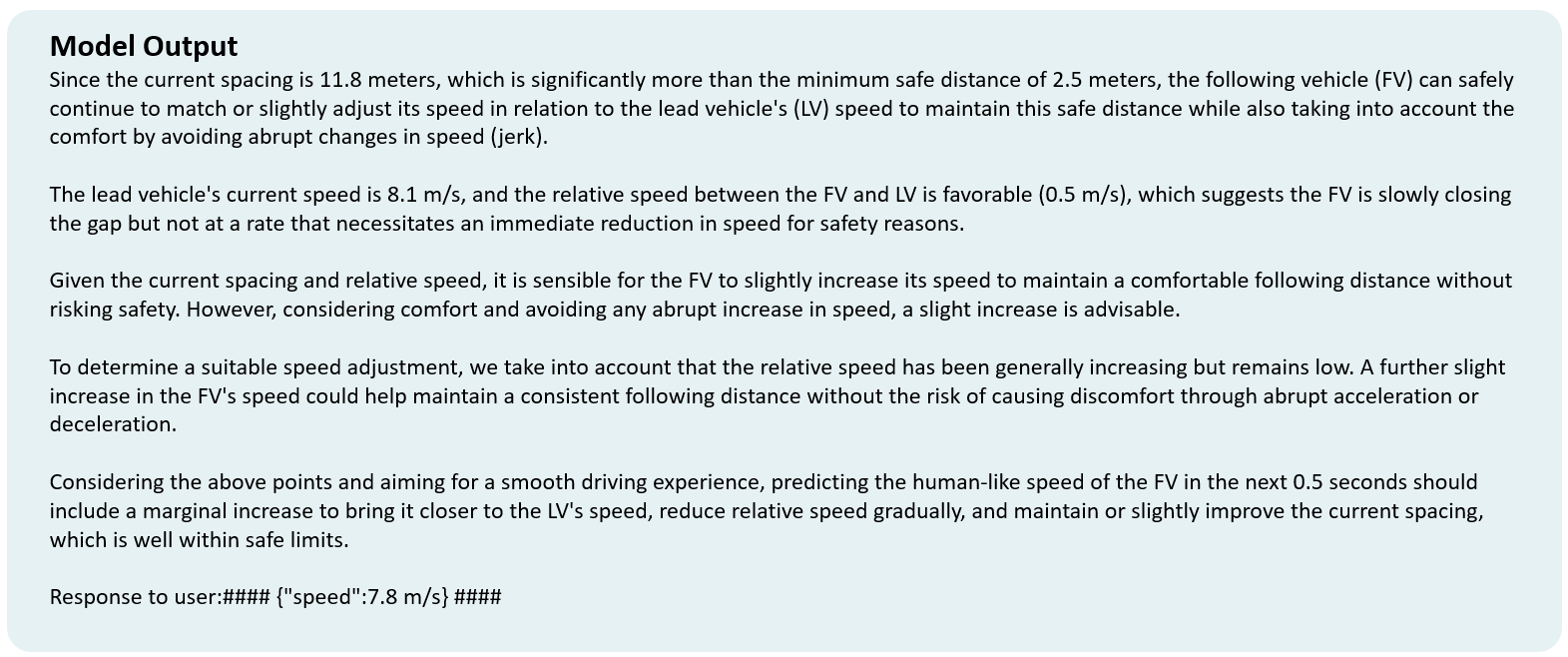}
    \caption{Detailed explanation of speed adjustment reasoning in car-following.}
    \label{fig:out}
\end{figure*}

Overall, our GenFollower model leverages the capabilities of LLMs to predict car-following behavior in autonomous driving systems. By formulating the prediction task as a language modeling problem and employing a zero-shot prompting approach, we can generate accurate and interpretable predictions for future trajectories and behaviors of the following vehicle. Figure~\ref{fig:out} serves as an illustrative example showcasing the detailed reasoning process and rationale behind the decision-making process for action.

% where $\mathcal{V}$ represents a linguistic depiction of the predicted speed, while $\mathcal{R}$ signifies a linguistic description of the cognitive chain of reasoning and decision-making process. $F_{GPT}$ denotes the GPT model, with $\mathcal{S}$ and $\mathcal{U}$ denoting the system message and user message, respectively. Unlike conventional motion planning methods focused solely on trajectory generation, our approach generates both speed predictions $\mathcal{V}$ and explicit reasoning processes $\mathcal{R}$, thereby enhancing transparency in decision-making processes. Consequently, our approach demonstrates superior interpretability compared to existing methodologies. Overall, our GenFollower model leverages the capabilities of LLMs to predict car-following behavior in autonomous driving systems. By formulating the prediction task as a language modeling problem and employing a zero-shot prompting approach, we can generate accurate and interpretable predictions for future trajectories and behaviors of the following vehicle. Figure~\ref{fig:out}  serves as an illustrative example showcasing the detailed reasoning process and rationale behind the decision-making process for action. 

% \begin{figure}[ht]
% \centering
% \includegraphics[width=0.8\linewidth]{car_following_scenario.png}
% \caption{Illustration of a car-following scenario, where the green vehicle represents the following vehicle (ego vehicle) and the blue vehicle represents the lead vehicle.}
% \label{fig:car_following_scenario}
% \end{figure}

\section{Data and Experiments}
% The experimental evaluation of the proposed GenFollower framework is conducted on the Waymo Open datasets.
\subsection{Waymo Open Dataset}
We evaluate the performance of the GenFollower framework on the Waymo Open Dataset~\cite{ettinger2021large}. This rich and diverse resource for autonomous vehicle research, offered by Waymo, provides high-resolution sensor data, including LiDAR, camera, and radar, captured from real-world driving environments. It includes detailed information on objects, their trajectories, and interactions, making it an invaluable tool for advancing machine learning models in perception, motion prediction, and planning. Its wide range of scenarios and comprehensive labeling facilitates robust development and benchmarking in the field of autonomous driving. Hu et al.~\cite{hu2022processing} and Chen et al.~\cite{chen2023follownet} extracted car-following events based on the Waymo Open dataset, which we leverage for our experiments. 

\subsection{Extracted Car-Following Data Description}

To ensure a representative sample size, we randomly selected 100 car-following events from our previous work~\cite{chen2023follownet} to evaluate the performance of the GenFollower. Each car-following event spans a duration of 15 seconds and comprises four key pieces of information captured at each time step:

\begin{itemize}
        \item \textbf{Spacing:} Represents the distance between the FV and the LV at each time step.
        \item \textbf{Following Vehicle Speed:} The speed of the FV at each time step.
        \item \textbf{Relative Speed:} Indicates the relative speed between the FV and the LV at each time step.
        \item \textbf{Lead Vehicle Speed:} The speed of the LV at each time step.
\end{itemize}

\subsection{Evaluation Metrics}

To evaluate the performance of the GenFollower model, we employ established metrics commonly used in car-following models. These metrics include:

\begin{enumerate}
%这里要说明 MSE 仅仅是和数据集中数据的接近程度，越小表明越接近Human Like Driving， 但是已经有许多研究表明，人类驾驶不一定是最好驾驶行为。
\item \textbf{Mean Squared Error (MSE):} The MSE between predicted and observed relative spacing is calculated to quantify the model's ability to accurately predict relative spacing. The MSE is computed using the formula:
\begin{equation}
MSE = \frac{1}{N} \sum_{i=1}^{N} \frac{1}{T} \sum_{j=1}^{T} [y_i(j) - \hat{y_i}(j)]^2
\end{equation}
where $N$ represents the number of testing events, $T$ denotes the time duration, $y_i(j)$ stands for the observed relative spacing, and $\hat{y_i}(j)$ represents the predicted relative spacing.
\item \textbf{Collision Rate:} The collision rate is another crucial metric used to evaluate the safety and reliability of trajectory generation. A collision is considered to occur in the simulated behavior if the relative spacing between two vehicles falls below 0. The formula for collision rate is as follows:
\begin{equation}
Collision\ Rate = \frac{\text{Number of events with spacing} < 0}{\text{Total number of car-following events}}
\end{equation}

\item \textbf{Time-To-Collision (TTC):} TTC represents the time it would take for the two vehicles to collide if they maintained their current states. A higher TTC value indicates a safer following distance between vehicles. The formula for calculating TTC is as follows:
\begin{equation} \label{eq:ttc}
TTC(t) = -\frac {S_{n-1, n}(t)}{ \Delta V_{n-1,n}(t)}
\end{equation}
where $S_{n-1, n}$ stands for the spacing distance between the FV and the LV, and $\Delta V_{n-1,n}$ is the relative speed (LV's speed $-$ FV's speed).
\end{enumerate}

\subsection{Baselines}

In this section, we compare the effectiveness of the proposed GenFollower model with five baseline models, including both physics-informed and data-driven approaches:

\begin{enumerate}
    \item \textbf{Gazis-Herman-Rothery (GHR)~\cite{gazis1961nonlinear}:} GHR is a physics-based safety model that determines the optimal acceleration of a following vehicle based on the driver's reaction time and sensitivity parameters. Parameter calibration is commonly performed using the Genetic Algorithm (GA)~\cite{zhu2018human}. 
    \item \textbf{Intelligent Driver Model (IDM):} IDM is another classic physics-based car-following model that considers various factors such as the driver's desired speed, desired time headway, maximum acceleration, comfortable deceleration, beta, and jam space. It measures the gap between the driver's expectations and the actual scenario. IDM is widely used in traffic simulation and control due to its simplicity and effectiveness. Similar to GHR, IDM is also calibrated using real-world datasets with GA.
    \item \textbf{Fully Connected Neural Network (NN):} This baseline model utilizes a fully connected neural network architecture, comprising multiple layers of interconnected nodes. The NN model's hyperparameters include the number of hidden layers and neurons in each layer, activation function, learning rate, optimizer, and batch size.
    \item \textbf{Long Short-Term Memory (LSTM) model:} LSTM, is a type of recurrent neural network designed to process sequences of data. Unlike the fully connected neural network, LSTM has several hyperparameters such as the number of LSTM layers, hidden units, and dropout probability.
    \item \textbf{Deep Deterministic Policy Gradient (DDPG)}: DDPG~\cite{zhu2018human} has been demonstrated effective in minimizing the discrepancy between simulated and actual actions to emulate human-like driving behaviors. Building upon this foundation, this study introduces DDPGs\_Max~\cite{chen2023follownet}, which incorporates a modified reward function featuring a max operation and collision penalty. This adaptation aims to reduce collisions and improve overall accuracy.
    % Specifically, the collision penalty encourages the agent to prioritize collision avoidance in the RL environment. Training terminates upon collision or upon completion of exploration of the entire car-following event.
    % During each simulation step, a max operation, inspired by max pooling, ensures comprehensive event exploration. The reward structure thus combines fidelity to observed values with incentives for collision-free operation.
\end{enumerate}

\begin{table*}[ht]
\caption{Results on the Waymo Open datasets.}
\label{result table}
    \centering
    \begin{tabular}{@{}lccc@{}}
    \toprule 
     & \textbf{\makecell{MSE of Spacing}} $\downarrow$ & \textbf{\makecell{Collision Rate  \% }} $\downarrow$  & \textbf{\makecell{Minimum TTC}} $\uparrow$ \\ 
     \midrule
        GHR & 50.22                                 & \textbf{0}             & 71.50 \\
        IDM & 37.25                                 & \textbf{0}             & 60.36\\
        NN  & \textbf{14.21}                       & 30                     & 33.69\\
        LSTM & 15.68                                & 2                      & 95.36\\
        DDPG & 35.85                             & \textbf{0}                     & \textbf{162.21}\\        
        % GenFollower (GPT-3.5)& 25.69                & 8                     & 74.31 \\      
        % GenFollower (GPT-3.5) with fine-tune& 22.44 & 6                      & 71.55 \\
        GenFollower (GPT-4) & 20.14                 & \textbf{0}             & 88.34 \\        
        \bottomrule
    \end{tabular}
\end{table*}

\subsection{Overall Performance}
Our evaluation on the Waymo Open Dataset reveals valuable insights into the performance of different car-following models. Table~\ref{result table} summarizes the results for MSE of Spacing, Collision Rate, and Minimum TTC.

\subsubsection{Mean Squared Error (MSE) of Spacing}

\begin{itemize}
    \item \textbf{NN Model}: Achieves the lowest MSE of Spacing at 14.21, indicating lowest average error in maintaining inter-vehicle distances.
    \item \textbf{GHR Model}: Shows the highest MSE of Spacing with 50.22, suggesting less precise control over spacing compared to other models.
    \item \textbf{GenFollower}: Compared to traditional models, our GenFollower achieves lower MSE in a zero-shot setting while ensuring collision-free performance. Our GenFollower demonstrates a balanced performance with zero collisions, competitive MSE, and a good Minimum TTC. This indicates its potential as a viable model for autonomous driving, combining safety with reasonable precision in spacing.
\end{itemize}

\subsubsection{Collision Rate}

% Importantly, several models achieved a rate of 0\%, including GHR, IDM, DDPG, and GenFollower (GPT-4). This highlights the effectiveness of these models in maintaining safe driving conditions.
\begin{itemize}
    \item \textbf{NN Model}: Exhibits the highest collision rate at 30\%, indicating a higher incidence of collisions in simulated scenarios.
    \item \textbf{GHR, IDM, DDPG, GenFollower (GPT-4) Models}: Achieve a  0\% collision rate, demonstrating the effectiveness of these models in maintaining safe driving conditions.
\end{itemize}

\subsubsection{Time-To-Collision (TTC)}

\begin{itemize}
    \item \textbf{DDPG Model}: Demonstrates exceptional performance with the longest minimum TTC of 162.21 seconds, offering an extensive warning time well ahead of potential collisions. This model excels in ensuring safety margins due to its crafted reward function.
    
    \item \textbf{NN Model}: Exhibits the shortest minimum TTC among the models evaluated, at 33.69 seconds. While providing a shorter warning time compared to others, it means more aggressive driving behavior.
    
    \item \textbf{GenFollower}: Shows a balance between maintaining safe distances and achieving reasonable reaction times, as evidenced by its consistently high minimum TTC values. This model leverages its predictive abilities to enhance collision avoidance strategies.
\end{itemize}

\subsubsection{Interpretability of GenFollower}

One key advantage of GenFollower is its ability to provide interpretable predictions through a step-by-step analysis process, as illustrated in Figure~\ref{fig:out}. Unlike black-box models, GenFollower leverages its extensive pre-training on vast datasets to systematically process and integrate complex information from the driving environment. This enables the generation of nuanced insights and well-founded responses based on a comprehensive understanding of the context.

% \textbf{Analysis Process:}
GenFollower's interpretability is facilitated by its step-by-step analysis process:
\begin{itemize}
    \item GenFollower iteratively analyzes input data, considering various contextual cues and patterns.
    \item Through this process, GenFollower builds up understanding and generates interpretable responses.
    \item Each step involves interpreting and synthesizing information to arrive at a coherent output.
\end{itemize}
This analysis allows humans to not only trust the model's predictions but also gain insights into the underlying reasoning process.

In contrast, for the GHR, IDM, NN, LSTM, and DDPG models, we followed the established training procedure outlined in FollowNet~\cite{chen2023follownet}, involving initial training on a designated training set and evaluation on a curated set of 100 test samples. Notably, our GenFollower (GPT-4) leveraged only prompt engineering techniques. This departure from conventional training paradigms underscores the capacity of large language models to achieve robust performance with either minimal or strategically chosen data inputs. Overall, GenFollower (GPT-4) demonstrates competitive performance with strong interpretability across all metrics, particularly in collision rate and TTC, underscoring its capability to perform effectively in autonomous driving scenarios without the traditional reliance on extensive training datasets.

\begin{table*}[ht]
\caption{Results of prompt engineering and fine-tuning on the Waymo Open datasets.}
\label{fine-tuning}
    \centering
    \begin{tabular}{@{}lccc@{}}
    \toprule 
     & \textbf{\makecell{MSE of Spacing}} $\downarrow$ & \textbf{\makecell{Collision Rate  \% }} $\downarrow$  & \textbf{\makecell{Minimum\_TTC}} $\uparrow$ \\ 
     \midrule        
        GenFollower (GPT-3.5)& 25.69                & 8                     & 74.31 \\      
        GenFollower (GPT-3.5) with fine-tune& 22.44 & 6                      & 71.55 \\
        GenFollower (GPT-4) & \textbf{20.14}                 & \textbf{0}             & \textbf{88.34} \\        
        \bottomrule
    \end{tabular}
\end{table*}
\subsection{Prompt Engineering vs. Fine-Tuning}
Table \ref{fine-tuning} presents a comparative analysis of prompt engineering and fine-tuning approaches using GenFollower models based on GPT-3.5 and GPT-4, evaluated on three aforementioned metrics. The GenFollower (GPT-3.5) model achieves an MSE of Spacing of 25.69, a Collision Rate of 8\%, and a TTC of 74.31. Fine-tuning the GenFollower (GPT-3.5) improves these metrics, reducing the MSE of Spacing to 22.44 and the Collision Rate to 6\%. In comparison, the GenFollower (GPT-4) model outperforms both GPT-3.5 variants, achieving the lowest MSE of Spacing at 20.14, a Collision Rate of 0\%, and the highest Minimum TTC of 88.34. These results demonstrate our GenFollower's superior overall performance in maintaining safe distances and minimizing collisions.

Overall, the GenFollower (GPT-3.5) demonstrates promise in achieving lower MSE even without explicit training on car-following data, particularly when prompted appropriately. Fine-tuning with human driver data further reduced MSE. Notably, the GenFollower (GPT-4) achieves even lower MSE in a zero-shot setting while ensuring collision-free performance.

\section{Summary and Conclusion}
Car-following behavior in traffic dynamics is critical for various applications, including traffic management and autonomous driving systems. In this study, we present GenFollower, a novel car-following model that leverages the power of large language models (LLMs) for improved trajectory prediction and analysis. By leveraging LLMs' ability to process sequential information and extract contextual cues, GenFollower reframes car-following as a language modeling problem and achieves competitive performance in both prediction accuracy and safety with the ability to provide explanations for its predictions on Waymo Open datasets. Notably, our GenFollower model employs a zero-shot approach, foregoing the need for fine-tuning. We harness its inherent common sense reasoning capabilities and self-explanation abilities to address the complexities of car-following behavior. 

In conclusion, GenFollower represents a pioneering application of LLMs specifically tailored for car-following tasks. It sets a new standard by integrating state-of-the-art language modeling techniques, offering a pathway toward more reliable and interpretable autonomous driving systems.

%总结全文, 未来的研究方向:由于大语言响应时间没有做到实时性的部署,部署到更加亲量化得模型；在更多的数据集测试大语言模型的泛化能力；由于大语言模型偶尔会生成无效的回答,设计更好的prompt以及安全筛选规格来优化；
% 实时性的考量 以及 错误得 以及泛化能力得测试

\section{Future Work}
Future research directions can build upon the foundation laid by GenFollower to further enhance the field of car-following modeling and autonomous driving systems.

\begin{itemize}
    \item Exploring ways to improve the robustness of GenFollower across diverse driving conditions and environments will be crucial for its practical applicability. This includes investigating adaptation techniques that allow the model to generalize effectively beyond training datasets.
    
    \item Deploying LLMs on more lightweight models to improve real-time deployment capabilities.
    
    \item Exploring the generalization capabilities of LLMs across diverse datasets will enhance their robustness and applicability..
    
    \item Developing better prompts and safety filtering mechanisms to mitigate the occasional generation of invalid responses by LLMs is crucial for improving overall reliability.
\end{itemize}
These efforts will contribute to making LLMs more efficient, reliable, and suitable for a wider range of practical applications in autonomous driving systems.

\bibliographystyle{IEEEtran}
\bibliography{reference}

%\newpage

\vspace{11pt}
%\bf{If you include a photo:}\vspace{-33pt}
\begin{IEEEbiography}[{\includegraphics[width=1in,height=1.25in,clip,keepaspectratio]{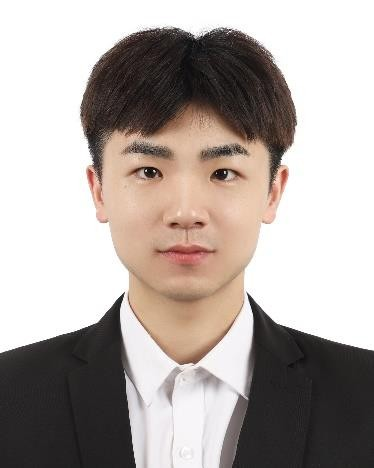}}]{Xianda Chen} received the M.S. degree from The University of Hong Kong, Hong Kong, China, in 2022. He is currently pursuing
a Ph.D. degree in intelligent transportation with The Hong Kong University of Science and Technology (Guangzhou), Guangzhou, China. His research interests include intelligent transportation, machine learning, and data analytics.
\end{IEEEbiography}

\begin{IEEEbiography}[{\includegraphics[width=1in,height=1.35in,clip,keepaspectratio]{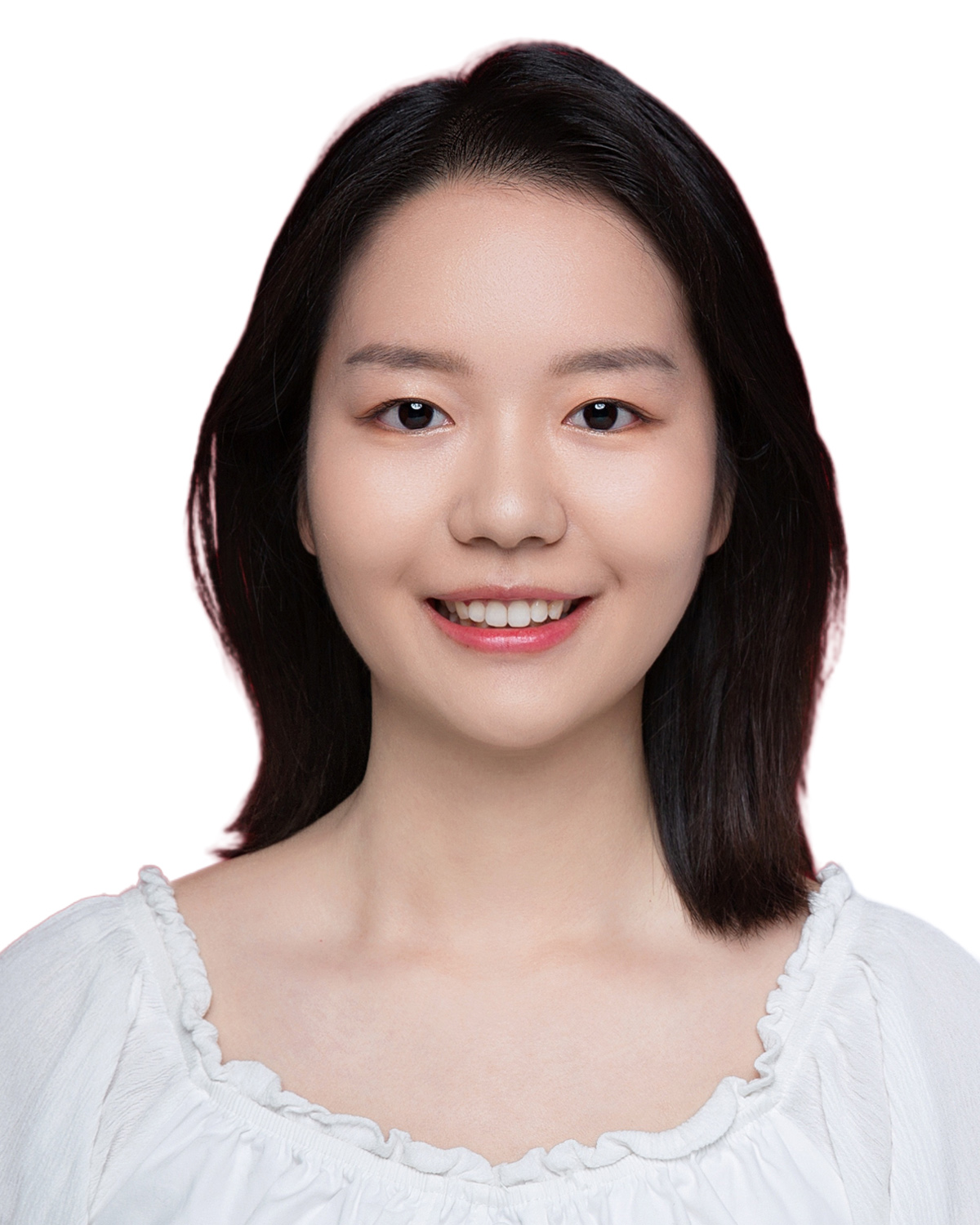}}]{Mingxing Peng} received a B.S. degree in digital media technology from Huaqiao University, China in 2018, and received a M.S. degree in computer technology from Sun Yat-Sen University in 2021. Currently, she is pursuing an Ph.D. degree under the Intelligent Transportation Thrust at the Systems Hub, the Hong Kong University of Science and Technology (Guangzhou). Her research interests include motion planning, trajectory prediction and large language models for autonomous driving.
\end{IEEEbiography}

\begin{IEEEbiography}
    [{\includegraphics[width=1in,height=1.25in,clip,keepaspectratio]{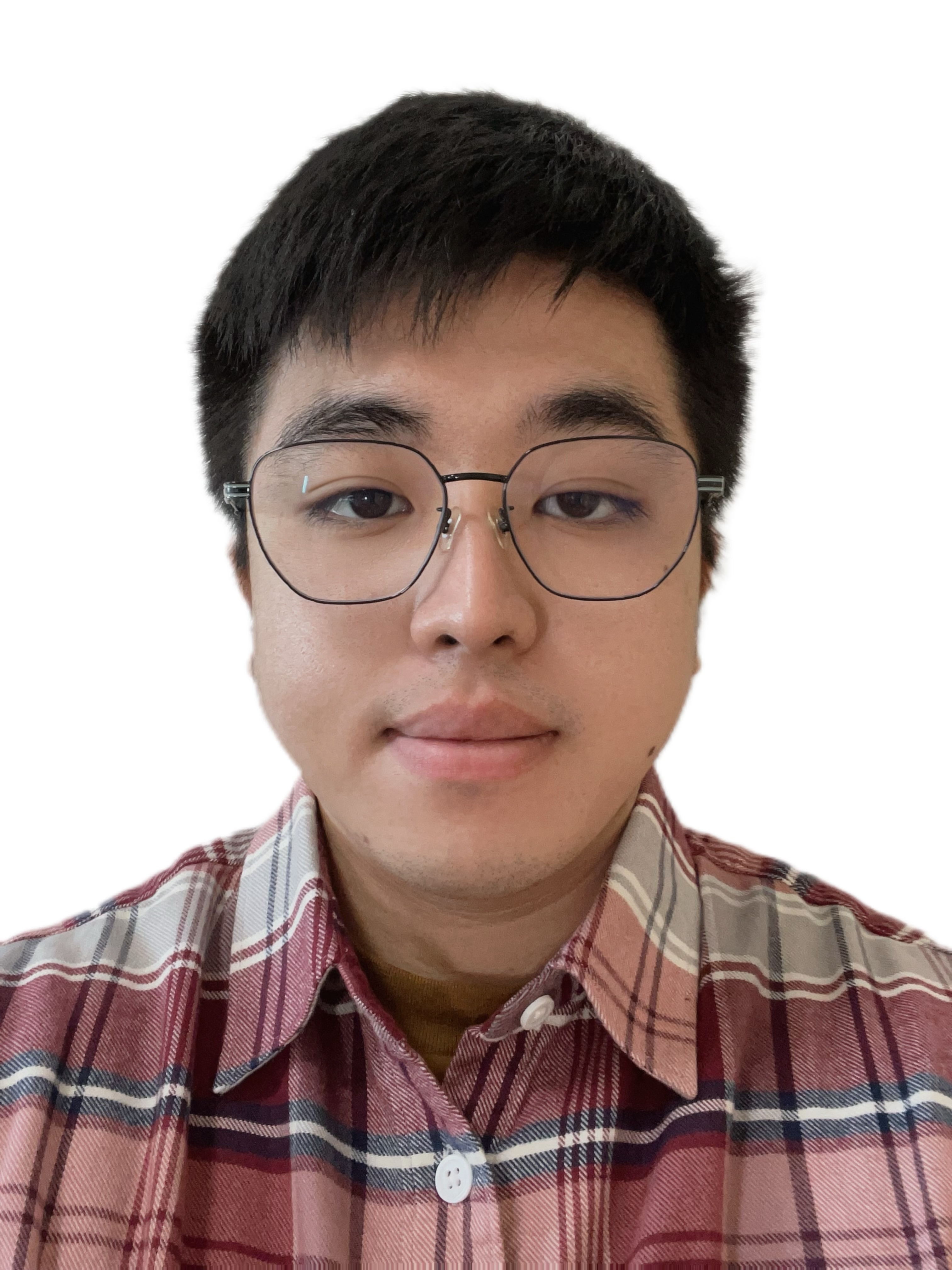}}]{PakHin Tiu}
    graduated with a Bachelor of Science degree in Applied Mathematics, Engineering, and Physics from the University of Wisconsin-Madison, United States, in 2020. He is currently pursuing a Master of Philosophy (MPhil) degree in Intelligent Transportation at the Hong Kong University of Science and Technology (Guangzhou), Guangzhou, China. His research interests encompass intelligent transportation and autonomous driving.
\end{IEEEbiography}

\begin{IEEEbiography}
    [{\includegraphics[width=1in,height=1.25in,clip,keepaspectratio]{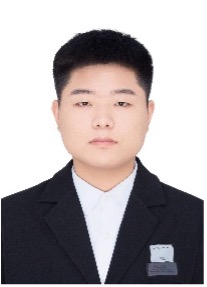}}]{Yuanfei Wu } is currently a research assistant at the Hong Kong University of Science and Technology (Guangzhou), China. His research interests include computer vision, intelligent transportation, and machine learning.
\end{IEEEbiography}

\begin{IEEEbiography}
    [{\includegraphics[width=1in,height=1.25in,clip,keepaspectratio]{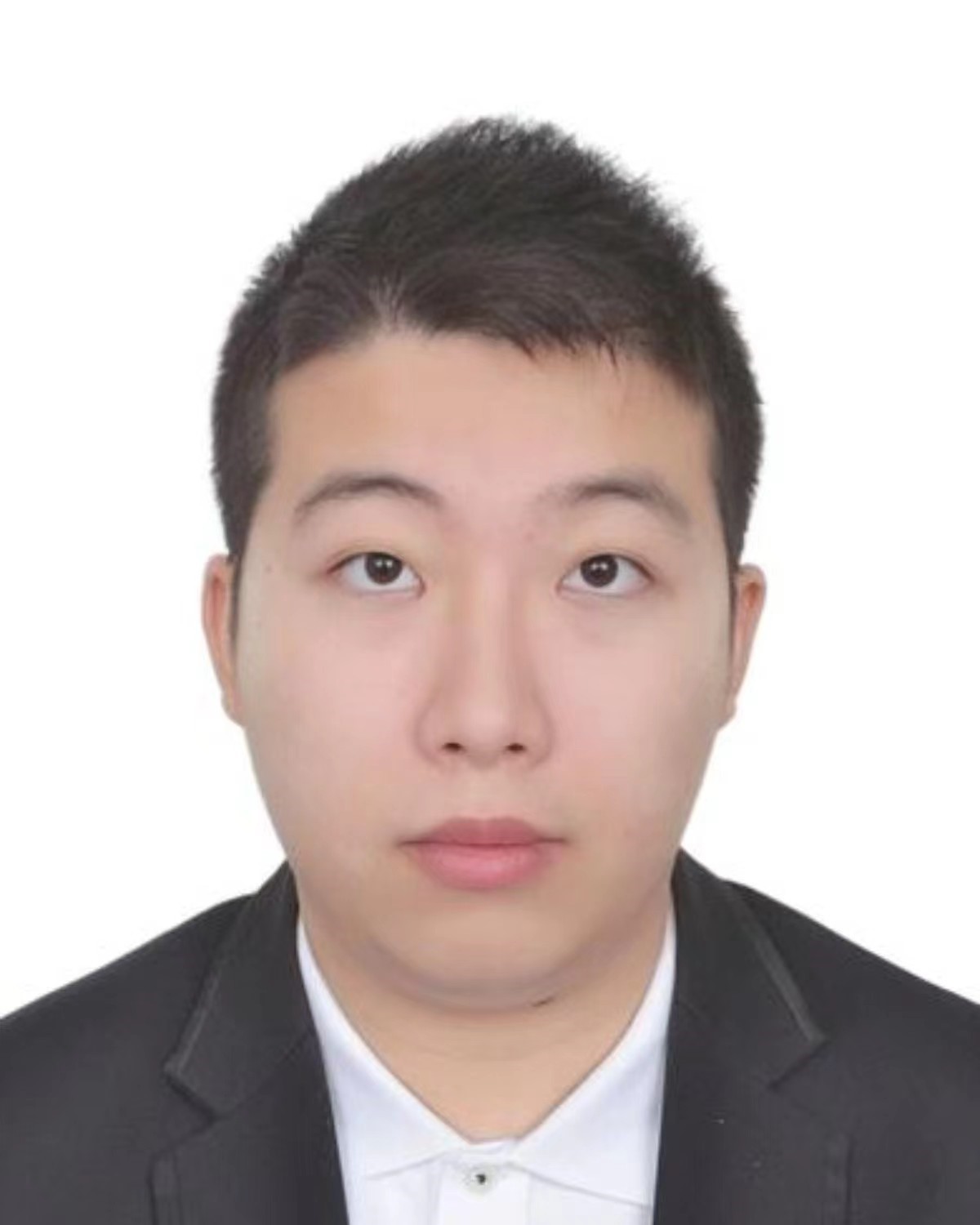}}]{Junjie Chen }
    received the B.SC degree from Beijing JiaoTong University, Beijing, China in 2023. He is currently pursuing a M.Phil degree in intelligent transportation with the Hong Kong University of Science and Technology (Guangzhou), China. His research interests include intelligent transportation, machine learning and traffic control.
\end{IEEEbiography}

\begin{IEEEbiography}[{\includegraphics[width=1in,height=1.25in,clip,keepaspectratio]{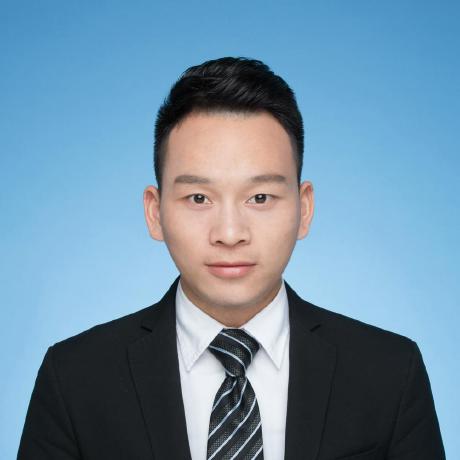}}]{Meixin Zhu} is a tenure-track Assistant Professor
in the Thrust of Intelligent Transportation (INTR)
under the Systems Hub at the Hong Kong University
of Science and Technology (Guangzhou). He is also
an affiliated Assistant Professor in the Civil and
Environmental Engineering Department at the Hong
Kong University of Science and Technology. He
obtained a Ph.D. degree in intelligent transportation
at the University of Washington (UW) in 2022. He
received his BS and MS degrees in traffic engineering in 2015 and 2018, respectively, from Tongji
University. His research interests include Autonomous Driving Decision
Making and Planning, Driving Behavior Modeling, Traffic-Flow Modeling
and Simulation, Traffic Signal Control, and (Multi-Agent) Reinforcement
Learning. He is a recipient of the TRB Best Dissertation Award (AED50)
in 2023.
\end{IEEEbiography}

\begin{IEEEbiography}[{\includegraphics[width=1in,height=1.25in,clip,keepaspectratio]{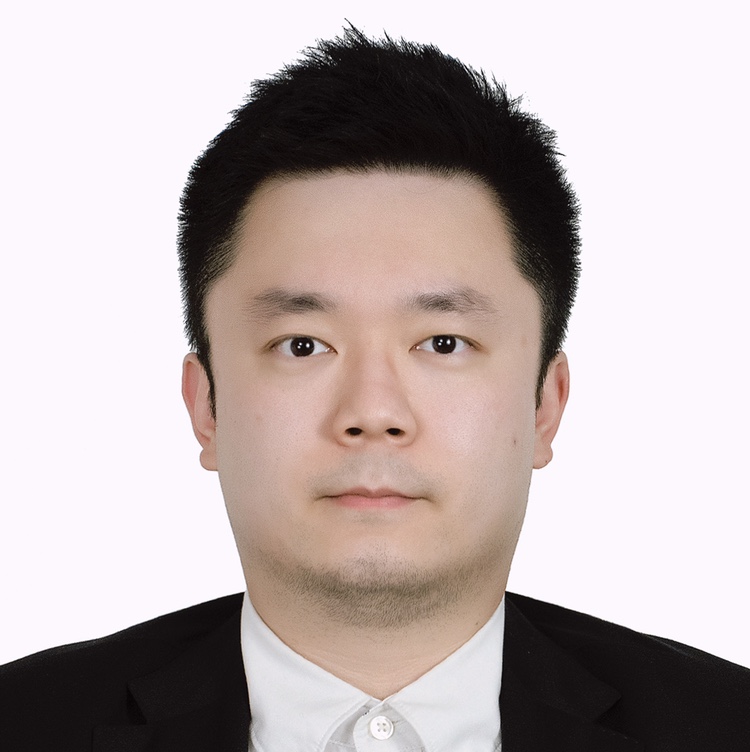}}]{Xinhu Zheng} received the Ph.D. degree in electrical and computer engineering from the University of
Minnesota, Minneapolis, in 2022. He is currently
an Assistant Professor with the Hong Kong University of Science and Technology (GZ). His current
research interests are data mining in power systems,
intelligent transportation system by exploiting different modality of data, leveraging optimization, and
machine learning techniques. 
\end{IEEEbiography}

\vfill

\end{document}